\documentclass[11pt,a4paper]{article}
\usepackage[T1]{fontenc}
\usepackage[utf8]{inputenc}
\usepackage{lmodern}
\usepackage[margin=2.5cm]{geometry}
\usepackage{microtype}
\usepackage{booktabs}
\usepackage{url}
\usepackage[numbers,sort&compress]{natbib}
\usepackage[hidelinks]{hyperref}
\usepackage{caption}
\newcommand{\doiPrereg}{10.5281/zenodo.23858965}
\newcommand{\doiRes}{10.5281/zenodo.23870450}
\newcommand{\tPub}{2026-11-01T13:52:48Z}
\newcommand{\tSperre}{2026-11-01T13:53:05Z}
\newcommand{\refDato}{2026-10-16}
\newcommand{\felles}{2025-01-01}
\newcommand{\cutBorealisTo}{2025-01-01}
\newcommand{\cutGemmaFire}{2025-01-31}
\newcommand{\revBTo}{6f14c8f8913a9d751b80eee1c4a72d2ca4bfaa98}

\newcommand{\startBTo}{2026-11-01T14:01:39Z}
\newcommand{\sluttBTo}{2026-11-01T15:51:50Z}
\newcommand{\revGFire}{4d7ae4984b7db7de8f8457170b3f1a419ee76d52}

\newcommand{\startGFire}{2026-11-01T16:00:01Z}
\newcommand{\sluttGFire}{2026-11-01T18:32:10Z}
\newcommand{\revBFireB}{939cf2ce2a5036151e719979f9b10fdc3d4b5ece}

\newcommand{\startBFireB}{2026-11-01T18:48:32Z}
\newcommand{\sluttBFireB}{2026-11-01T19:56:14Z}
\newcommand{\revBOpen}{9445eb2acee3dd7adf7124781f06962ccd47c870}

\newcommand{\startBOpen}{2026-11-01T19:59:21Z}
\newcommand{\sluttBOpen}{2026-11-01T20:55:13Z}
\newcommand{\vllm}{0.21.0}
\newcommand{\transformers}{5.8.1}
\newcommand{\gpu}{A100-80GB}
\newcommand{\maksTok}{1,024}

\newcommand{\temp}{0.0}
\newcommand{\seedGen}{20261117}
\newcommand{\bit}{60}
\newcommand{\bootB}{10,000}
\newcommand{\seed}{20261117}
\newcommand{\alfa}{0.05}
\newcommand{\sesoi}{10}
\newcommand{\lnRegelFra}{168}
\newcommand{\lnRegelTil}{170}
\newcommand{\mddUten}{3.32}
\newcommand{\mddMed}{4.42}
\newcommand{\vToTomme}{349}
\newcommand{\vToMaks}{16,384}
\newcommand{\simulaSide}{11}
\newcommand{\simulaPdfSide}{12}
\newcommand{\simulaDato}{25 June 2026}
\newcommand{\simulaSitat}{Part of the advantage is parametric knowledge of Norwegian rules and thresholds, which drift as the rules change; the rubric scores confident correct recall as a pass, yet the same behaviour becomes the confident-hallucination failure mode 3 (§4.1) once a recalled fact is stale, so the comparative ranking is a snapshot.}
\newcommand{\prismaN}{90}

\newcommand{\otsBlokk}{974179}

\newcommand{\nAlle}{916}

\newcommand{\nPrim}{503}

\newcommand{\nSerier}{89}
\newcommand{\Hn}{411}
\newcommand{\Hbti}{10}
\newcommand{\Hbnull}{11}
\newcommand{\Hd}{\ensuremath{-}0.2}
\newcommand{\Hlo}{\ensuremath{-}2.7}
\newcommand{\Hhi}{+2.5}
\newcommand{\Hp}{0.668}
\newcommand{\Htilstand}{no effect of interest}

\newcommand{\SToBti}{4}
\newcommand{\SToBnull}{31}
\newcommand{\SToD}{\ensuremath{-}5.4}
\newcommand{\SToP}{\ensuremath{<}\,0.0001}
\newcommand{\nTestbare}{5}
\newcommand{\bN}{411}
\newcommand{\bDisk}{0}

\newcommand{\islBTo}{34}
\newcommand{\plassBTo}{21}
\newcommand{\annenBTo}{0}
\newcommand{\ingenBTo}{13}

\newcommand{\lengdeBTo}{1}
\newcommand{\islGFire}{58}
\newcommand{\plassGFire}{17}
\newcommand{\annenGFire}{0}
\newcommand{\ingenGFire}{41}

\newcommand{\islBFireB}{159}
\newcommand{\plassBFireB}{1}
\newcommand{\annenBFireB}{148}
\newcommand{\ingenBFireB}{10}
\newcommand{\ulikeBFireB}{264}
\newcommand{\toppEnBFireB}{217}
\newcommand{\toppToBFireB}{138}

\newcommand{\islBOpen}{5}
\newcommand{\plassBOpen}{0}
\newcommand{\annenBOpen}{5}
\newcommand{\ingenBOpen}{0}
\newcommand{\ulikeBOpen}{41}
\newcommand{\toppEnBOpen}{222}
\newcommand{\toppToBOpen}{184}

\newcommand{\bortFelles}{17}
\newcommand{\kildeangivelse}{Inneholder data under Norsk lisens for offentlige data (NLOD) tilgjengeliggjort av Stiftelsen Lovdata. (Contains data under the Norwegian Licence for Open Government Data (NLOD) made available by Stiftelsen Lovdata.) Licence: NLOD 2.0, \url{https://data.norge.no/nlod/no/2.0}. Source: \url{https://api.lovdata.no/}. The items were derived from these data by norregel (extraction, chaining, question templates); these changes were made by the author and are not Lovdata's responsibility.}

\newcommand{\orcid}[1]{\href{https://orcid.org/#1}{#1}}

\title{Outdated numeric values from Norwegian statutes and regulations in a further-trained model and its base model: a preregistered paired comparison of Borealis 2 and Gemma 4}
\author{Eirik Botten Nicolaysen\\EcoDeco AS\\ORCID \orcid{0009-0001-9188-6788}\\eirik@ecodeco.no}
\date{}

\begin{document}
\maketitle

\begin{abstract}
We compare how often two released language models give outdated numeric values from Norwegian statutes and regulations:
NbAiLab/borealis2-26b-a4b-preview (Borealis 2), which the National Library of Norway further trained from
google/gemma-4-26B-A4B-it (Gemma 4), and Gemma 4 itself. The study was preregistered (doi:\doiPrereg) and uses the
closed norregel benchmark, scored deterministically against a value history. On the \Hn{} of \nPrim{} primary items where
both models gave an interpretable answer, Borealis 2 alone was outdated on \Hbti{} items and Gemma 4 alone on \Hbnull{}
($d = \Hd$ percentage points, 95\% cluster bootstrap interval $[\Hlo, \Hhi]$, exact one-sided McNemar $p = \Hp$). By the
preregistered rule this is \emph{\Htilstand}: the whole interval lies within the smallest effect of interest of
$\pm\sesoi$ percentage points. Among the secondary measures, only the share of answers giving a value that took effect after
the question date differs after Holm correction; Gemma 4 gave such answers more often. An exploratory comparison of two smaller Borealis
models had no discordant pairs and is not informative. The design compares released models and does not isolate the
effect of any part of their training.
\end{abstract}

\section{Introduction}

The Simula CAISS audit of Borealis~1, which is built on Gemma~3, points to outdated legal values without measuring
them \citep{simula2026}. On printed p.~\simulaSide{} (PDF page~\simulaPdfSide) of the report dated \simulaDato{} it states:
``\simulaSitat'' The present study measures one part of this for Borealis~2: on the same dated questions, does the
further-trained model give outdated values more often than the model it is built on?

The study is descriptive. Borealis~2 and Gemma~4 differ in the National Library's continued pre-training and
fine-tuning, of which newspaper text is one of many changes, and each model comes from a single training run. A
difference between them could therefore not be attributed to newspaper text or to any other part of the training
(preregistration §2, §14).

\section{Related work}

The preregistration reports a scoping search that included \prismaN{} works; its limits (a criterion added after the
search had started, a stopping rule that was not met, and candidates that were not screened) are stated there
\citep{prereg}. The five closest works are FiscalQA~Pro, French tax law with deterministic scoring and numeric tolerance
\citep{fiscalqa}; Prior, Schultz and Grabmair, German law, with the failure modes post-cutoff staleness and recency bias,
scored by an LLM judge \citep{prior2026}; Feng, numeric state-policy values with a same-state stale error class
\citep{feng2026}; TIDE, the value in force on a date, for tariff rates \citep{tide2026}; and evolveQA, evolving knowledge
\citep{evolveqa}. This study differs from these by using Norwegian statutes and regulations from Lovdata, deterministic
scoring against a value history \citep{norregelarxiv}, and a paired comparison of a nationally further-trained model
with its base model. It makes no claim to be the first such comparison.

\section{Method}

\paragraph{Preregistration.} The study was registered on Zenodo (doi:\doiPrereg, published \tPub). No norregel question
was run on the models before the registration; the embargo was lifted at \tSperre{} after the publication was confirmed.
The analysis code, its lock file, the run script and the item manifest were frozen in the preregistration.

\paragraph{Hypotheses.} H-a (confirmatory): among interpretable answers, the share of outdated answers is higher for
Borealis~2 than for Gemma~4 on the same items. H-b (exploratory): the share is higher for NbAiLab/borealis-4b than for
NbAiLab/borealis-open-4b, two Borealis~1 models whose fine-tuning data differ, according to the model card, by tasks
derived from newspaper material \citep{cardborealisfourb}. H-b enters the confirmatory analysis only as secondary measure
S-5.

\paragraph{Models and run.} Table~\ref{tab:models} lists the four models with their pinned revisions
\citep{cardborealistwo,cardgemmafour,cardborealisfourb,cardborealisopen}. Each model ran on Modal on one \gpu{} GPU in
bf16 with vLLM~\vllm{} and transformers~\transformers, greedy decoding (temperature \temp, seed \seedGen, which has no
effect at temperature \temp), a maximum of \maksTok{} new tokens, reasoning off, and the question as the only user
message, without a system prompt. Each container checked the revision, the vLLM version and the chat template before
answering. Answers were written as a hash chain in chunks of at most \bit{} questions; the chain heads were stamped with
OpenTimestamps. Borealis~2 ran from \startBTo{} to \sluttBTo, Gemma~4 from \startGFire{} to \sluttGFire, borealis-4b
from \startBFireB{} to \sluttBFireB{} and borealis-open-4b from \startBOpen{} to \sluttBOpen.

\begin{table}[t]\centering\small
\caption{Models and pinned revisions.}\label{tab:models}
\begin{tabular}{lll}
\toprule
Role & Hugging Face ID & Revision \\
\midrule
(a) & \texttt{NbAiLab/borealis2-26b-a4b-preview} & \texttt{\revBTo} \\
(a) & \texttt{google/gemma-4-26B-A4B-it} & \texttt{\revGFire} \\
(b) & \texttt{NbAiLab/borealis-4b} & \texttt{\revBFireB} \\
(b) & \texttt{NbAiLab/borealis-open-4b} & \texttt{\revBOpen} \\
\bottomrule
\end{tabular}
\end{table}

\paragraph{Items and reference date.} The items, the answer key and the scorer come from norregel
\citep{norregelconcept,norregelvtwo,norregelarxiv}: \nAlle{} dated numeric values from Lovdata's open data, with
reference date \refDato. The items are closed; the manifest holds item IDs only. The primary items are the H1 set of
norregel v2 whose answer-key value took effect on or before \felles, the earlier of the two models' stated training
cutoffs (Borealis~2 \cutBorealisTo, Gemma~4 \cutGemmaFire): \nPrim{} items in \nSerier{} series (Table~\ref{tab:flow}).

\paragraph{Scoring and analysis.} Answers are classed by norregel's locked scorer v2 as correct, outdated (the answer equals an
earlier value in the item's series but not the answer key), other error, or uninterpreted; uninterpreted is an outcome, never counted as an
error. The primary measure is the share of outdated answers among interpretable answers, paired on the items where both
models' answers are interpretable. H-a is tested with an exact one-sided McNemar test \citep{mcnemar1947},
$\alpha = \alfa$, with $d = (b_{10} - b_{01})/n$ and a 95\% cluster bootstrap interval over series ($B = \bootB$, seed
\seed). The decision rule is set in the preregistration (lines \lnRegelFra--\lnRegelTil): \emph{supported} if
$p < \alfa$ and the lower bound is above zero; \emph{no effect of interest} if the whole interval lies within
$\pm\sesoi$ percentage points, the smallest effect of interest taken over from norregel v2; otherwise
\emph{inconclusive}. The planned minimum detectable difference was \mddMed{} percentage points with cluster adjustment
(\mddUten{} without), from a proxy pair in the v2 data. Secondary measures with at least one discordant pair are
corrected with Holm's procedure \citep{holm1979}.

\section{Results}

\begin{table}[t]\centering\small
\caption{Item flow from all items to the analysed pairs.}\label{tab:flow}
\begin{tabular}{lrrp{6.4cm}}
\toprule
Step & Items & Dropped & Reason for dropping \\
\midrule
All items, run on every model & 916 & -- & -- \\
Primary set & 880 & 36 & in norregel's \texttt{excluded\_items}, as in v2 \\
H1 set within the primary set & 567 & 313 & not in norregel v2's \texttt{h1\_eligible\_items} \\
Primary items (a) and (b) & 503 & 64 & key took effect after 2025-01-01 \\
Pairs, both interpretable, (a) & 411 & 92 & uninterpreted: Borealis 2 only 57, Gemma 4 only 33, both 2 \\
Pairs, both interpretable, (b) & 411 & 92 & uninterpreted: borealis-4b only 92, borealis-open-4b only 0, both 0 \\
\bottomrule
\end{tabular}
\end{table}

\paragraph{Completeness.} All four models answered all \nAlle{} items; every chain is intact, with one revision and one
vLLM version per model. Table~\ref{tab:flow} shows how the \nAlle{} items narrow to the analysed pairs. In both pairs
\Hn{} items remain, but the items that drop out differ: \bortFelles{} of them are the same in both pairs.

\paragraph{H-a (confirmatory).} Of the \Hn{} pairs, Borealis~2 alone was outdated on $b_{10} = \Hbti$ items and Gemma~4
alone on $b_{01} = \Hbnull$. The difference is $d = \Hd$ percentage points, 95\% cluster bootstrap interval
$[\Hlo, \Hhi]$, and the exact one-sided McNemar test gives $p = \Hp$. The $p$-value is not below $\alfa$, the lower bound is
not above zero, and the whole interval lies within $\pm\sesoi$ percentage points. By the preregistered rule the result is
\emph{\Htilstand}. No other reading of the interval against the smallest effect of interest is made.

\paragraph{Per model.} Table~\ref{tab:classes} gives the classes on the primary items for each model, unpaired.

\begin{table}[t]\centering\small
\caption{Classes on the 503 primary items per model (unpaired). The share of outdated answers uses each model's interpretable answers as the denominator.}\label{tab:classes}
\begin{tabular}{lrrrrr}
\toprule
Model & Correct & Outdated & Other error & Uninterpreted & Outdated / interpretable \\
\midrule
Borealis 2 & 14 & 25 & 405 & 59 & 25 / 444 (5.6\%) \\
Gemma 4 & 23 & 26 & 419 & 35 & 26 / 468 (5.6\%) \\
borealis-4b & 2 & 6 & 403 & 92 & 6 / 411 (1.5\%) \\
borealis-open-4b & 2 & 6 & 495 & 0 & 6 / 503 (1.2\%) \\
\bottomrule
\end{tabular}
\end{table}

\paragraph{Secondary measures.} Table~\ref{tab:secondary} gives the secondary measures. After Holm correction over the
\nTestbare{} testable measures, the null hypothesis is rejected only for S-2 (later value: a value that took effect after
the question date): Gemma~4 alone gave such a value on \SToBnull{} items
and Borealis~2 alone on \SToBti{} ($d = \SToD$ percentage points, two-sided $p \SToP$). The study does not identify a
cause for this difference.

\begin{table}[t]\centering\small
\caption{Secondary measures, paired on the primary items. Two-sided exact McNemar; Holm over the 5 measures with at least one discordant pair. $b_{10}$: first model only; $b_{01}$: second model only. S-5 compares borealis-4b with borealis-open-4b; the others Borealis 2 with Gemma 4.}\label{tab:secondary}
\begin{tabular}{lrrrrrl}
\toprule
Measure & $n$ & $b_{10}$ & $b_{01}$ & $d$ (pp) & $p$ & Holm \\
\midrule
S-1 outdated, all answers & 503 & 10 & 11 & \ensuremath{-}0.2 & 1.0000 & not rejected \\
S-2 later value & 503 & 4 & 31 & \ensuremath{-}5.4 & \ensuremath{<}\,0.0001 & rejected \\
S-3 correct & 503 & 4 & 13 & \ensuremath{-}1.8 & 0.0490 & not rejected \\
S-4 uninterpreted & 503 & 57 & 33 & +4.8 & 0.0149 & not rejected \\
S-6 outdated = most recent differing value & 411 & 2 & 10 & \ensuremath{-}1.9 & 0.0386 & not rejected \\
S-5 (b) outdated, both interpretable & 411 & 0 & 0 & +0.0 & -- & not testable \\
\bottomrule
\end{tabular}
\end{table}

\paragraph{Uninterpreted answers.} Table~\ref{tab:unint} gives the uninterpreted answers. ``No answer line'' means that
the locked scorer found no answer line of the required form, which requires the label at the start of a line. Over all
\nAlle{} items, Borealis~2 has \islBTo{} such answers: \plassBTo{} end with the template placeholder \texttt{SVAR:
<verdi>} or an empty \texttt{SVAR:}, \annenBTo{} contain \texttt{SVAR:} in another position, and \ingenBTo{} contain no
\texttt{SVAR:}. Gemma~4 has \islGFire{} (\plassGFire, \annenGFire{} and \ingenGFire), borealis-4b \islBFireB{}
(\plassBFireB, \annenBFireB{} and \ingenBFireB) and borealis-open-4b \islBOpen{} (\plassBOpen, \annenBOpen{} and
\ingenBOpen). The placeholder answers are missing answers; the answers with \texttt{SVAR:} in another position are a
format outcome of the locked scorer. Neither group is re-scored.

\begin{table}[t]\centering\small
\caption{Uninterpreted answers per model, on the primary items and on all items, with subclasses.}\label{tab:unint}
\begin{tabular}{lrp{3.9cm}rp{3.9cm}}
\toprule
Model & Primary & Subclasses & All 916 & Subclasses \\
\midrule
Borealis 2 & 59 & refusal 49, no answer line 9, several numbers 1 & 157 & refusal 119, no answer line 34, several numbers 4 \\
Gemma 4 & 35 & no answer line 26, refusal 9 & 131 & refusal 73, no answer line 58 \\
borealis-4b & 92 & no answer line 92 & 159 & no answer line 159 \\
borealis-open-4b & 0 & -- & 5 & no answer line 5 \\
\bottomrule
\end{tabular}
\end{table}

\paragraph{(b), exploratory.} Within the registered analysis, H-b enters as S-5, which has no discordant pairs and
cannot be tested (Table~\ref{tab:secondary}). An exploratory copy of the analysis, with only the test label changed,
gives the same picture for the primary measure: $n = \bN$ and \bDisk{} discordant pairs. Both models repeated
their output heavily: borealis-4b \ulikeBFireB{} distinct completions over \nAlle{} items, the two most frequent
occurring \toppEnBFireB{} and \toppToBFireB{} times, and borealis-open-4b \ulikeBOpen{} distinct completions, the two
most frequent occurring \toppEnBOpen{} and \toppToBOpen{} times. With no discordant pairs and such output, (b) is not
informative about H-b. Whether the training of the two models was otherwise identical is not known.

\section{Limitations}

The design does not isolate newspaper text or any other part of the training: (a) compares two released models that
differ in continued pre-training and fine-tuning, each from one training run. Borealis~2 is a preview model. Greedy
decoding on a GPU does not guarantee identical answers across runs. The maximum of \maksTok{} tokens is lower than the
\vToMaks{} used for API models in v2; \lengdeBTo{} Borealis~2 answer stopped at the maximum. Some answers in the ``no answer
line'' subclass are template placeholders, that is, missing answers, and they reduce the number of analysed pairs. An
outdated answer cannot be told apart from a correct value given for the wrong date, and the answer key has residual
errors (norregel v2). The training cutoffs are those stated by the providers; for Borealis~2 no separate cutoff is
stated for its fine-tuning data or its newspaper data. The Borealis~2 licence text calls the model a derivative of
Gemma~3, while its base is Gemma~4. The planned minimum detectable difference rests on a proxy pair from the v2 data and
was computed for a two-sided test with a normal approximation, whereas the primary test is exact and one-sided; the
power of the exact one-sided test was not computed.

\section{Relation to norregel v2}

Gemma~4 was measured in norregel v2 through Google's API, with thinking on and a maximum of \vToMaks{} tokens; \vToTomme{}
of its \nAlle{} answers there were empty because they stopped at the maximum. Here it ran locally in bf16 with thinking off
and a maximum of \maksTok{} tokens. The numbers here are therefore not a replication of v2, and differences between v2
and this study are not interpreted.

\section{Deviations}

The deviations during the run are listed in the result report \citep{resultrecord}; the full deviation log with
timestamps is kept in the author's register and is not deposited. In short: D-1, the operator script that drove
the chunks labels a stop at the cost cap ``rule~7'', whereas the preregistration places the cost cap under rule~6; no
stop occurred. D-2, the (b) models were run with a second copy of the run script, which checks the Gemma~3 chat format
taken from \texttt{chat\_template.jinja} at the pinned revisions instead of the model card, after the first copy failed
the template check before any norregel question was run. D-3, the exploratory (b) analysis uses a copy of the analysis
code with only the test label changed. D-4, the cost cap was set before the run and enforced before every chunk, but was
entered in the deviation log only after the run. D-5, during the run the hash chains were held only in a non-persistent
location until the author's instruction to copy them after each model; the copies are byte-identical with the analysed
chains. D-6 to D-8 concern cost bookkeeping, the descriptive scripts and Modal's run history. None changes a locked file. Of the deviations recorded before the run, those that bear on it (A-20 to A-24 in the
author's register) are described in the preregistration (§3, §8, §15).

\section{Data and code availability}

The preregistration with the frozen files is at doi:\doiPrereg{} \citep{prereg}. The results record \citep{resultrecord}, DOI \doiRes, holds the result report, the analysis outputs, the item
flow, the descriptive supplement, the four chain heads with their OpenTimestamps stamp (confirmed in Bitcoin block
\otsBlokk) and the code that is not frozen: the operator script, the (b) run-script copies, the (b) analysis copy and the
descriptive and report scripts. The frozen run script and analysis code are in the preregistration. Raw model answers
are not published. The NB-License of Borealis~2 and borealis-4b ties every use of model output to its use restrictions,
and the Gemma Terms of Use of borealis-open-4b make the user responsible for outputs and their use; the Gemma~4 answers
(Apache~2.0) are withheld as well, because the pairs are analysed together. The items are closed. The norregel code, the tests that run without the items and the published documentation are
also at \url{https://github.com/avalyset/norregel}.

\paragraph{Source attribution.} \kildeangivelse{} \citep{nlod}

\section*{AI disclosure}

Claude (Anthropic), including Claude Code, was used as a tool under the author's direction for study-design support, the literature search, drafting the preregistration, the report and this paper, operating the model runs and running the locked analysis; independent Claude subagents checked the documents. The author is responsible for the content.

\section*{Competing interests}

The author is the founder and CEO of EcoDeco AS, which offers independent evaluation of language models. EcoDeco AS
received no payment for this study.

\section*{Funding}

This work received no external funding.

\section*{Acknowledgements}

GPU compute was paid with Modal's free monthly credits (Starter plan).

\bibliographystyle{plainnat}
\bibliography{refs}

\end{document}